\documentclass[conference]{IEEEtran}
\IEEEoverridecommandlockouts

\usepackage{cite}
\usepackage{amsmath,amssymb,amsfonts}
\usepackage{graphicx}
\usepackage{textcomp}

\usepackage{bm}
\usepackage{hyperref}
\usepackage{url}
\usepackage{epstopdf}
\usepackage{enumitem}
\usepackage{algorithm}
\usepackage{algpseudocode}
\usepackage{amsmath, amsthm}
\usepackage{booktabs}
\usepackage{multirow}

\usepackage{xcolor}

\theoremstyle{plain}
\newtheorem{theorem}{Theorem}
\newtheorem{lemma}[theorem]{Lemma}
\newtheorem{proposition}[theorem]{Proposition}

\theoremstyle{definition}
\newtheorem{definition}[theorem]{Definition}

\newtheorem{assumption}{Assumption}

\theoremstyle{remark}

\def\BibTeX{{\rm B\kern-.05em{\sc i\kern-.025em b}\kern-.08em
    T\kern-.1667em\lower.7ex\hbox{E}\kern-.125emX}}

\begin{document}

\title{Overcoming Spurious Correlations in Conditional Diffusion Models via Doubly Robust Target Correction
}

\author{\IEEEauthorblockN{Eiichi Sakurai}
\IEEEauthorblockA{
\textit{Graduate School of Information Science and Technology,} \\
\textit{The University of Tokyo, }\\
Tokyo, Japan \\
\\
\textit{Artificial Intelligence Research Center,}\\
\textit{National Institute of Advanced Industrial Science and Technology,}\\
Tokyo, Japan \\
e.sakurai@aist.go.jp
\vspace{-35pt}
}\\

\and
\IEEEauthorblockN{Kenji Yamanishi}
\IEEEauthorblockA{\textit{Graduate School of Information Science and Technology
,} \\
\textit{The University of Tokyo, }\\
Tokyo, Japan \\
yamanishi@g.ecc.u-tokyo.ac.jp}
}

\maketitle

\begingroup
\renewcommand{\thefootnote}{}
\footnotetext{
This paper presents a comprehensive version of the work
submitted to IEEE BigData 2026. The work was accepted
for publication as a short paper. Due to the page
limitations of the conference format, some details,
analyses, and experimental results from the original
full-length manuscript were omitted from the accepted
short paper. This arXiv version provides the complete
presentation of the work, including these additional
materials.

\vspace{0.5em} 
\noindent
\copyright 2026 IEEE. Personal use of this material is permitted. Permission from IEEE must be obtained for all other uses, in any current or future media, including reprinting/republishing this material for advertising or promotional purposes, creating new collective works, for resale or redistribution to servers or lists, or reuse of any copyrighted component of this work in other works.
}
\endgroup

\begin{abstract}
While conditional diffusion models have achieved remarkable success in 
generation tasks, they memorize spurious correlations within training 
data. This causes them to deviate from ideal target distributions. 
Minimizing a covariate shift-adjusted target risk can address this issue. 
However, retraining models from scratch for different distributions is 
computationally prohibitive. To overcome this limitation, we propose 
DR-Diff, a framework that efficiently fine-tunes pre-trained conditional 
diffusion models. Specifically, DR-Diff leverages a functional gradient 
approach to derive a pseudo-target. This pseudo-target captures the 
residual between the base model's noise prediction and the ground-truth 
noise under the target distribution. Training the model to fit this 
pseudo-target enables efficient target adaptation. Theoretically, we 
prove that DR-Diff guarantees target risk reduction at the population level. 
It also achieves a 
pseudo-doubly robust property, providing resilience against density 
ratio estimation errors and base model misspecifications. Empirically, 
DR-Diff matches the noise prediction accuracy of models trained from 
scratch on synthetic datasets. Furthermore, it mitigates inter-group 
disparities in downstream classification tasks. 
\end{abstract}

\begin{IEEEkeywords}
out of distribution, debiasing, Diffusion Model
\end{IEEEkeywords}


\section{Introduction.}
Diffusion models have demonstrated remarkable success in conditional 
generation, enabling high-fidelity synthesis of target variables $y_0$ 
under conditions $(X_{\rm label}, X_{\rm cond})$ 
consisting of a class label and auxiliary attributes
\cite{10.5555/3495724.3496298, ho2021classifierfree}. By iteratively 
denoising corrupted samples, these models effectively learn the 
observational conditional distribution $P(y_0 | X_{\rm label}, X_{\rm cond})$, 
thereby offering a flexible generative mechanism. 

However, in real-world datasets, conditioning variables $X_{\rm label}$ 
and $X_{\rm cond}$ often exhibit strong spurious dependencies due to 
observational biases. Because standard conditional training minimizes 
the evidence lower bound over the empirical distribution, it memorizes 
these spurious correlations. This issue is particularly evident in 
out-of-distribution (OOD) regions, where specific combinations of 
$(X_{\rm label}, X_{\rm cond})$ rarely appear, causing the generator to 
synthesize distorted samples. Overcoming these spurious correlations 
essentially requires adapting the model to a covariate-shifted target 
distribution. 


To address this covariate shift toward a user-specified target 
distribution, methods like importance weighting 
\cite{NEURIPS2024_4d254c96,10.1145/3711896.3736819} and causal 
interventions \cite{sanchez22adiffscm,pmlr-v202-de-sousa-ribeiro23a} 
have been explored. However, these approaches typically require 
retraining models from scratch for each target shift, making them 
computationally prohibitive. Bridging this distribution gap efficiently 
without retraining is our fundamental objective. 


Although addressing this covariate shift is essential, retraining 
diffusion models from scratch for each target distribution is 
computationally prohibitive. To overcome this challenge, we propose 
DR-Diff, an efficient fine-tuning framework based on functional 
gradients. 

Our main contributions are summarized as follows: 
\begin{itemize}[leftmargin=*, topsep=0pt, itemsep=0pt]
 \item \textbf{Robust Diffusion Framework via Functional Gradients:} By 
       replacing the standard noise target with a derived pseudo-target 
       that captures the residual between base model predictions and the 
       true noise, our framework efficiently adapts base models into 
       accurate noise prediction functions under the desired target 
       distribution through fine-tuning.  
 \item \textbf{Theoretical Guarantees and Pseudo-DR Property:} We 
       theoretically demonstrate that, at the population level, our 
       formulation guarantees a reduction in the expected noise 
       prediction error given an appropriate step size $\eta$. 
       Furthermore, we characterize the pseudo-Doubly Robust property of 
       our pseudo-target, illustrating how it provides resilience 
       against both density ratio estimation errors and base model 
       misspecifications. 
 \item \textbf{Empirical Effectiveness for Covariate-Shift Adaptation:} 
       Evaluations on synthetic datasets demonstrate that our method 
       achieves noise prediction performance comparable to models 
       trained from scratch under the target distribution. Moreover, 
       validations on real-world datasets (LSAC and Jigsaw) show that 
       our method synthesizes target-aligned samples that improve the 
       robustness of downstream classifiers against spurious 
       correlations by reducing inter-group disparities (e.g., 
       differences in False Positive Rates). 
\end{itemize}

\section{Related Works.}
\subsection{Standard Conditional Generation with Diffusion Models}
Following the success of diffusion models \cite{10.5555/3495724.3496298}, 
research on conditional generative models has significantly advanced. 
Initially, classifier-guided diffusion models 
\cite{NEURIPS2021_49ad23d1} were developed, but CFG~\cite{ho2021classifierfree} has since emerged and become the 
standard, particularly in image generation. 
Concurrently, inpainting-based approaches have also been explored 
as a form of conditional generation,
where unmasked regions serve as the condition for generating the image
 (e.g., RePaint \cite{lugmayr2022repaint}, CoPaint\cite{zhang2023COPAINT}). 
However, because these methods approximate the joint 
data distribution between condition and output directly from training 
data, any spurious correlations present in the training data inevitably 
bias the generated outputs. 

\subsection{Bias Mitigation in Generative Models.}
To overcome such biases in the training data, there are two primary 
directions. The first integrates diffusion models with causal graphs 
that describe the data generation process, such as Diff-SCM 
\cite{sanchez22adiffscm},DSCM\cite{pmlr-v202-de-sousa-ribeiro23a} 
and SPATIAL \cite{pmlr-v267-rasal25a}. While 
causally rigorous, these methods require a fully specified causal graph, 
which is often intractable for complex or high-dimensional real-world 
datasets. The second approach addresses covariate shifts by modifying 
the training objective, using techniques like propensity-score weighting 
(e.g., Diff-PO \cite{NEURIPS2024_4d254c96}, DIME\cite{10.1145/3711896.3736819}) or 
Bayesian perturbation (e.g. \cite{doi:10.1126/sciadv.ads4593}). 
Although loss-level interventions avoid the need for 
explicit causal graphs, existing methods require retraining the entire 
diffusion model from scratch for every target distribution.  

Our proposed framework, DR-Diff, aligns with this loss-intervention 
approach but overcomes its prohibitive computational burden by learning 
the target noise prediction via residual fine-tuning from a pre-trained 
base model.

\section{Proposed Method.}
\subsection{Problem Formulation and Structural Dependencies.}
\label{problemsetting}
We consider a data-generating process with triples ($y_0$, $X_{\rm label}$, 
$X_{\rm cond}$), where $y_0 \in \mathbb{R}^d$ is the generated output, 
$X_{\rm label}$ is the primary condition (e.g., class label), and $X_{\rm cond}$ 
represents auxiliary attributes. In practice, $X_{\rm label}$ and $X_{\rm cond}$ 
often exhibit complex structural dependencies rather than being 
independent. We assume the following underlying structure: 
\vspace{-5pt}
\[ 
     X_{\rm cond} \to X_{\rm label}, \quad X_{\rm cond} \to y_0, \quad X_{\rm label} \to y_0.
 \vspace{-5pt}
\]
Under this structure, standard conditional diffusion models learn the 
biased observational distribution $P(y_0 | X_{\rm label}, X_{\rm cond})$. Our 
goal is \textit{debiased generation}: estimating the \textbf{debiased  
generative distribution} by isolating the direct 
relationship of each condition to the output and removing the spurious 
coupling between $X_{\rm label}$ and $X_{\rm cond}$. This ensures the model 
faithfully extrapolates the effect of the primary label even in rare, 
OOD scenarios. 

To guarantee identifiability and establish a foundation for the Neyman 
orthogonality discussed later, we posit the following standard 
assumption. 

\begin{assumption}[Sufficiency of Observed Attributes]
The auxiliary attributes $X_{\rm cond}$ are sufficient to account for all 
 structural biases between $X_{\rm label}$ and $y_0$. We assume that no 
 unobserved factors jointly influence these variables, thus ensuring the 
 validity of subsequent propensity estimation. 
\end{assumption}
\begin{assumption}[Compactness and Positivity]
\label{assumption_positivity}
The joint condition space $\mathcal{X}$ of $X = (X_{\rm label}, X_{\rm cond})$ 
is assumed to be a compact domain (e.g., a finite product of compact 
intervals for continuous variables and finite discrete sets for 
categorical variables). Furthermore, for all 
$X \in \mathcal{X}$, 
there exists a strictly positive probability of observing 
 them jointly under the biased observational distribution: $P(X_{\rm label} 
 \mid X_{\rm cond}) > 0$. 
\end{assumption}
\subsection{Preliminaries: Conditional Diffusion Models.}
Diffusion models generate data by reversing a forward process that 
incrementally adds Gaussian noise to output $y_0$ 
\cite{10.5555/3495724.3496298}.
The state at time $t$ can be expressed as 
$y_t = \sqrt{\bar{\alpha}_t} y_0 + \sqrt{1 - \bar{\alpha}_t} \epsilon$, 
where $\epsilon \sim \mathcal{N}(0, I)$ and $\bar{\alpha}_t$ 
follows a predefined noise schedule.
 Training typically simplifies the variational lower bound 
of the noise prediction task as follows:
\[
 \mathcal{L}_{\rm simple} = \mathbb{E}_{t, y_0, \epsilon} \left[ || \epsilon - \epsilon_\theta(y_t, t) ||^2 \right],
\]
where $\epsilon_\theta$ is a neural network trained to estimate 
the added noise $\epsilon$ from $y_t$ and $t \sim \mathcal{U}(1, T)$.

For conditional generation, let $X = (X_{\rm label}, X_{\rm cond})$ denote 
the combined conditioning vector. 
The denoising network is then conditioned on $X$, formulated as 
$\epsilon_\theta(y_t, t, X)$.
The objective is the expectation over the empirical distribution $p_{\rm emp}$:
\[
 \mathcal{L}_{\rm cond} = \mathbb{E}_{X \sim p_{\rm emp}(X),y_0 \sim p(y_0|X),t, \epsilon} 
\left[ || \epsilon - \epsilon_\theta(y_t, t, X) ||^2 \right].
\]

The objective of the generative model is not to imitate the empirical 
distribution $p_{\rm emp}(X)$ of the training data, but to perform accurate 
conditional generation under a target distribution $p_{\rm target}(X) = 
p(X_{\rm label})p(X_{\rm cond})$, representing a scenario where each condition 
can be controlled independently. To achieve this, the loss function must 
be defined as an expectation over $p_{\rm target}(X)$. However, we cannot 
directly evaluate this target loss from the training data due to the 
covariate shift. Therefore, how to indirectly optimize this $\mathcal{L}_{\rm target}$ 
using the available empirical distribution becomes the main challenge 
addressed in the subsequent sections. 
\subsection{Weight-based Bias Correction and Its Limitations}
The most straightforward approach
to address the covariate shift between $p_{\rm emp}(X)$ and $p_{\rm target}(X)$
is to incorporate importance weighting into the loss function. 
Indeed, as a prior approach to mitigating 
spurious dependencies between conditioning variables in diffusion models, 
instance reweighting of the objective function has been proposed 
\cite{NEURIPS2024_4d254c96, doi:10.1126/sciadv.ads4593}. Under this 
approach, the loss function is described as: 
\[
 \mathcal{L}_{\rm weighted}(\epsilon_\theta,w) = 
 \mathbb{E}_{t, (y_0,X)\sim P_{\rm emp}, \epsilon} 
 \left[ w(X) || \epsilon - \epsilon_\theta
 ||^2 \right].
\]

For example, Diff-PO \cite{NEURIPS2024_4d254c96} defines the weight $w(X)$ as 
the Inverse Probability Weight based on the relationship 
between conditioning variables and variables corresponding to confounders.
 In our generalized setting targeting arbitrary covariate shifts, the 
 optimal weight function corresponds to the likelihood ratio 
$w_0(X) = p_{\text{\rm target}}(X) / p_{\text{\rm emp}}(X)$. 

As shown in Diff-PO, this weighted objective satisfies the Neyman 
orthogonality condition defined in \cite{OrthogonalStatisticalLearning}. We show 
that this property holds even under our generalized likelihood ratio 
$w(X)$. 

\begin{proposition}\label{prop_neyman_orthogonality}
 Under Assumptions 1 and 2, $\mathcal{L}_{\text{\rm weighted}}(\epsilon_\theta, w)$ 
 satisfies Neyman Orthogonality with respect to the 
 estimation errors $w$. 
\end{proposition}

The proof is provided in Appendix \ref{proof:neyman_orthogonality}. 
Owing to this property, theoretically minimizing 
this objective allows the target model to learn at a 
sufficient convergence rate, even if the nuisance estimator $w(X)$ 
contains initial biases.

However, this theoretical guarantee relies on 
the assumption that the true noise prediction is realizable within the 
given neural network model. 
For example, it is known that noise prediction is often difficult for 
standard diffusion models on data like the Swiss Roll, and it is  
expected that weighting alone would be insufficient for such data.

To achieve robust learning that can withstand such limitations, we propose 
an alternative Doubly Robust approach in the next section. 

\subsection{Doubly Robust Diffusion (DR-Diff)}
Minimizing $\mathcal{L}_{\text{\rm weighted}}(\epsilon_\theta, w_0)$ theoretically 
yields the optimal noise predictor for a target distribution. However, 
retraining diffusion models from scratch is computationally prohibitive. 
With pre-trained models now widely available, fine-tuning them for the 
target distribution is far more desirable. To fulfill this need, we propose a 
novel framework that learns an optimal perturbation to a pre-trained 
base model. Specifically, we utilize the functional gradient of 
$\mathcal{L}_{\text{\rm weighted}}(\epsilon_\theta, w_0)$ to derive a pseudo-target 
that directly guides model refinement. 

Let $\hat{\epsilon}_{\text{\rm base}}$ be the base model to be updated. 
Since the weighted risk 
$\frac{1}{2}\mathcal{L}_{\text{\rm weighted}}(\epsilon_\theta, w_0)$ is a 
functional with respect to the function $\epsilon_\theta$, its 
functional gradient is defined by the directional derivative. For a 
small variation $s \cdot\delta \epsilon$ in the function space, the 
derivative is calculated as follows: 
\begin{align}
& \left. \partial_s  \mathcal{L}_{\text{\rm weighted}}(\hat{\epsilon}_{\text{\rm base}} + s \delta \epsilon, w_0)/2\> \right|_{s=0} 
\nonumber\\
&= \mathbb{E}_{P_{\text{\rm emp}}, \epsilon, t} \left[ -w_0(X) (\epsilon - \hat{\epsilon}_{\text{\rm base}})^T \delta \epsilon \right]
\end{align}%
Therefore, the functional gradient $\delta \mathcal{L}(y_t, t, X)$ at 
the base model $\hat{\epsilon}_{\text{\rm base}}$ is $-w_0(X)(\epsilon - 
\hat{\epsilon}_{\text{\rm base}})$. The operation of updating the function 
in the direction of this gradient with a step size $-\eta$, namely 
$\hat{\epsilon}_{\text{\rm base}} - \eta \delta \mathcal{L}$, works to 
improve the evaluation of the base model under the target distribution.  

Since the exact gradient calculation depends on the unobservable 
ground-truth noise $\epsilon$, a direct functional addition for the 
update is intractable. Moreover, because the true density ratio $w_0(X)$ 
remains unknown, it is necessary to replace it with an empirical 
estimator $\hat{w}(X)$. 
Therefore, starting from $\hat{\epsilon}_{\text{\rm base}}$, training 
$\epsilon_\theta$ to predict the updated noise 
$\hat{\epsilon}_{\text{\rm base}} - \eta \delta 
\mathcal{L}(\hat{\epsilon}_{\text{\rm base}}, \hat{w}(X))$ is exactly 
equivalent to minimizing the squared error against a pseudo-target 
$\tilde{\epsilon}_{\text{\rm DR}}$, defined as follows: 
\begin{align}
\tilde{\epsilon}_{\text{\rm DR}} &= \hat{\epsilon}_{\text{\rm base}}(y_t, t, X) + \eta \hat{w}(X) \left( \epsilon - \hat{\epsilon}_{\text{\rm base}}(y_t, t, X) \right), \\
\mathcal{L}_{\text{\rm DR}} &= \mathbb{E}_{t, \epsilon, (y_0,X)\sim P_\text{\rm emp}} \left[ ||\epsilon_\theta(y_t, t, X) - \tilde{\epsilon}_{\text{\rm DR}}||^2 \right].
\end{align}
From this formulation, we show that learning this pseudo-target under an 
appropriate step size $\eta$ decreases the loss under the 
target distribution $\mathcal{L}_{\text{\rm weighted}}(\epsilon_\theta, 
w_0)$. 

\begin{theorem}[ Guarantee of Target Risk Reduction]
\label{theo_riskreduction}
Let the updated function be defined as 
$\epsilon_{\text{\rm DR}}:= \mathbb{E}[\tilde{\epsilon}_{\text{\rm DR}} \mid y_t, t, X]$. 
If the step size satisfies $\eta < 2/\max \hat{w}(X)$, then the following 
 inequality holds: 
\vspace{-5pt}
\begin{equation}
\vspace{-5pt}
\mathcal{L}_{\text{\rm weighted}}(\hat{\epsilon}_{\text{\rm base}}, w_0) \ge 
 \mathcal{L}_{\text{\rm weighted}}(\epsilon_{\text{\rm DR}}, w_0). 
\end{equation}
\end{theorem}

\begin{proof}
Let the true ideal function be $\epsilon^* = \mathbb{E}[\epsilon \mid 
 y_t, t, X]$ and set $\delta = \hat{w}(X)(\epsilon^* - 
 \hat{\epsilon}_{\text{\rm base}})$. The expected value of the updated 
 function can be written as $\epsilon_{\text{\rm DR}} = 
 \hat{\epsilon}_{\text{\rm base}} + \eta \delta$. 
Calculating the difference in the expected squared 
 error under the target distribution before and after the update yields 
 the following: 
\begin{align*}
&( \mathbb{E}_{P_{\text{\rm target}}, \epsilon, t} \left[ ||\epsilon - \epsilon_{\text{\rm DR}}||^2 \right] -  \mathbb{E}_{P_{\text{\rm target}}, \epsilon, t} \left[ ||\epsilon - \hat{\epsilon}_{\text{\rm base}}||^2 \right] )/2\\
&= \mathbb{E} \left[ \left( -\eta \hat{w}(X) + \eta^2/{2} \hat{w}^2(X) \right) ||\epsilon^* - \hat{\epsilon}_{\text{\rm base}}||^2 \right].
\end{align*}
Therefore, as long as $0 < \eta \hat{w}(X) < 2$, the loss decreases. 
\end{proof}

This theorem demonstrates that the condition for loss reduction depends 
only on the maximum value of $\hat{w}(X)$, indicating that the 
estimation error of the density ratio model has minimal impact on this 
guarantee. 
Furthermore, as shown in the proof above, the expected value after the 
update is $\mathbb{E}[\tilde{\epsilon}_{\text{\rm DR}} \mid y_t, t, X] = 
\hat{\epsilon}_{\text{\rm base}} + \eta \hat{w}(X) (\epsilon^* - 
\hat{\epsilon}_{\text{\rm base}})$. Thus, it is also possible to 
algebraically recover the true noise function $\epsilon^*$ from the 
optimized function $\mathbb{E}[\tilde{\epsilon}_{\text{\rm DR}} \mid y_t, t, 
X]$
\footnote{We do not directly apply equation (\ref{noiserecover}) 
during inference due to numerical instability caused by 
the inverse weighting $1/\hat{w}(X)$. 
Instead, we use the trained model $\epsilon_\theta$.
While Theorem \ref{theo_riskreduction} guarantees target risk reduction 
at the population level, 
practical performance depends on the approximation and optimization 
errors of $\epsilon_{\theta}$ bounded during finite-sample training. 
} 
. 

\begin{lemma}[Recovery of the Target Denoising Function]
Assuming $\hat{w}(X) > 0$ for all valid $X$ and $\eta \neq 0$, the true 
 target denoising function $\epsilon^*$ can be algebraically recovered 
 from the optimal model $\epsilon_{\text{\rm DR}}$ as follows: 
\vspace{-5pt}
\begin{equation}
\label{noiserecover}
\epsilon^* = \hat{\epsilon}_{\text{\rm base}} + 
 \left( \epsilon_{\text{\rm DR}} - \hat{\epsilon}_{\text{\rm base}} \right) 
/(\eta \hat{w}(X))
\end{equation}
\end{lemma}

\subsection{Connection to Doubly Robust Estimation}
A Doubly Robust (DR) estimator, recognized in causal inference 
and off-policy evaluation, takes the following form: 
\vspace{-5pt}
\[
\vspace{-5pt}
\begin{aligned}
 &\text{DR Estimator} \nonumber\\
 &= \text{Base Model} + \text{Weight} \times (\text{Observation} - \text{Base Model})\nonumber
\end{aligned}%
\]

When $\eta=1$, our proposed pseudo-target 
$\tilde{\epsilon}_{\text{\rm DR}} = \hat{\epsilon}_{\text{\rm base}} + 
\eta \hat{w}(X) (\epsilon - \hat{\epsilon}_{\text{\rm base}})$ 
shares this mathematical structure. 
Because this formulation is conditioned on each point 
$(y_t, t, X)$, 
DR-Diff exhibits a ``pseudo-Doubly Robust'' property from the 
following two perspectives: 

\begin{enumerate}
    \item \textbf{Robustness to Density Ratio Estimation Error:} Even if 
	  the density ratio model $\hat{w}(X)$ is inaccurate, if the 
	  base model $\hat{\epsilon}_{\text{\rm base}}$ perfectly matches 
	  the true function $\epsilon^*$, then 
	  $\tilde{\epsilon}_{\text{\rm DR}} = \epsilon^*$. Consequently, the 
	  error in $\hat{w}(X)$ does not adversely affect the noise 
	  prediction.  
    \item \textbf{Robustness to Base Model Misspecification:} Conversely, 
	  even if the base model $\hat{\epsilon}_{\text{\rm base}}$ is 
	  inaccurate, as long as the density ratio $\hat{w}(X)$ is 
	  correctly estimated, the proper gradient direction toward 
	  $\epsilon^*$ at $\hat{\epsilon}_{\text{\rm base}}$ is obtained, 
	  guaranteeing a correct model update. 
\end{enumerate}
\subsection{Practical Algorithm and Nuisance Estimation}
Building upon the theoretical debiasing properties shown in the previous 
sections, our practical training algorithm directly optimizes the 
DR-corrected target $\tilde{\epsilon}_{\text{\rm DR}}$, as outlined in 
Algorithm \ref{alg:dr_finetuning}. The nuisance functions required 
within this process are estimated as follows: 

\subsubsection{Estimation of Nuisance $\hat{w}$ via Cross-fitting.}
The likelihood ratio $\hat{w}(X)$ relies on the conditional distribution 
$\hat{P}(X_{\text{\rm label}} \mid X_{\text{\rm cond}})$. Therefore, using 
predictions from a model trained on the entire dataset induces 
own-observation bias, which distorts the final estimator. To circumvent 
this problem, we employ $K$-fold cross-fitting \cite{DML}. The dataset is 
partitioned into $K$ disjoint subsets, and for any data point in $D_i$, 
its density ratio $\hat{w}$ is computed using a nuisance model trained 
on the complement $D \setminus D_i$. 

\subsubsection{Pre-training the Baseline $\hat{\epsilon}_{\text{\rm base}}$.}
To suppress the variance of the DR estimator, the baseline predictor 
$\hat{\epsilon}_{\text{\rm base}}$ requires high-fidelity noise prediction 
capabilities. For this baseline, we can employ a conditional diffusion 
model pre-trained on the entire observed dataset, using either a 
standard empirical objective or a weighted objective. Since 
the prediction target here is randomly sampled noise $\epsilon$, 
own-observation bias does not occur during the baseline prediction, 
allowing us to safely use the full dataset without $K$-fold 
cross-fitting. 
\begin{center}
\begin{algorithm}[tb] 
\caption{Proposed Method: DR-Diff}
\label{alg:dr_finetuning}
\begin{algorithmic}[1]
\Require Observed dataset $\mathcal{D}$, Pre-trained baseline predictor 
 $\hat{\epsilon}_{\rm base}$,  Density ratio estimator $\hat{w}$, step size 
 $\eta$, learning rate $\gamma$.
\State Initialize main model $\epsilon_\theta \leftarrow \hat{\epsilon}_{\rm base}$. 
\Repeat
    \State Sample $(y_0, X) \sim \mathcal{D}.$ 
    \State Sample $t \sim \text{Uniform}(\{1, \dots, T\})$.
    \State Sample $\epsilon \sim \mathcal{N}(0, \mathbf{I})$.
    \State Compute noisy output $y_t = \sqrt{\bar{\alpha}_t} y_0 + \sqrt{1 - \bar{\alpha}_t} \epsilon$.
    \State Compute baseline prediction $e = \hat{\epsilon}_{\rm base}(y_t, t, X)$.
    \State Compute density ratio $w = \hat{w}(X_{\rm label},X_{\rm cond})=\hat{P}(X_{\rm label})/\hat{P}(X_{\rm label}|X_{\rm cond})$.
    \State Construct pseudo-target  $\tilde{\epsilon}_{\rm DR} = e + \eta\cdot w\cdot (\epsilon - e).$
    \State Parameter update by gradient descent: 
    $\theta \leftarrow \theta - \gamma\nabla_\theta \left\| \epsilon_\theta(y_t, t, X) - \tilde{\epsilon}_{\rm DR} \right\|^2$.
\Until convergence
\Ensure Debiased main model $\epsilon_\theta$.
\end{algorithmic}
\end{algorithm}
\vspace{-10pt}
\end{center}

\section{Experiments.}
In this section, we empirically evaluate the proposed DR-Diff framework. 
First, using synthetic datasets, we verify its capability to generate 
data accurately even under severe OOD scenarios. 
Next, we demonstrate its practical utility in mitigating structural 
biases to train fair classifiers on real-world datasets. 

\subsection{Validation on Synthetic Datasets.}\label{synthetic_exp}
\subsubsection{Objective and Experimental Setup}
To evaluate the generative capabilities, we designed two synthetic 
datasets featuring strong correlations among the generation conditions. 
We trained our models on these datasets and verified their ability to 
generate data in regions where observations are sparse due to these 
biases. The data generation mechanisms are as follows: 
\begin{enumerate}
   \item \textbf{Causal Cluster:} 
         A model where the generation center on $y_0 \in \mathbb{R}^2$ shifts 
         depending on $X_{\rm label}$ and $X_{\rm cond}$.
   \item \textbf{Swiss Roll (10D):} 
         A complex model where the progression and angular velocity 
         of the spiral are determined by $X_{\rm label}$ and $X_{\rm cond} \in \mathbb{R}^2$,
         which is then projected into a 10-dimensional space via a random matrix 
         for observation.
\end{enumerate}
Detailed formulations are provided in the Appendix \ref{syntheticgen}.

\begin{figure}[tb]
\vspace{-5pt}
\begin{center}
\includegraphics[width=1.0\linewidth]{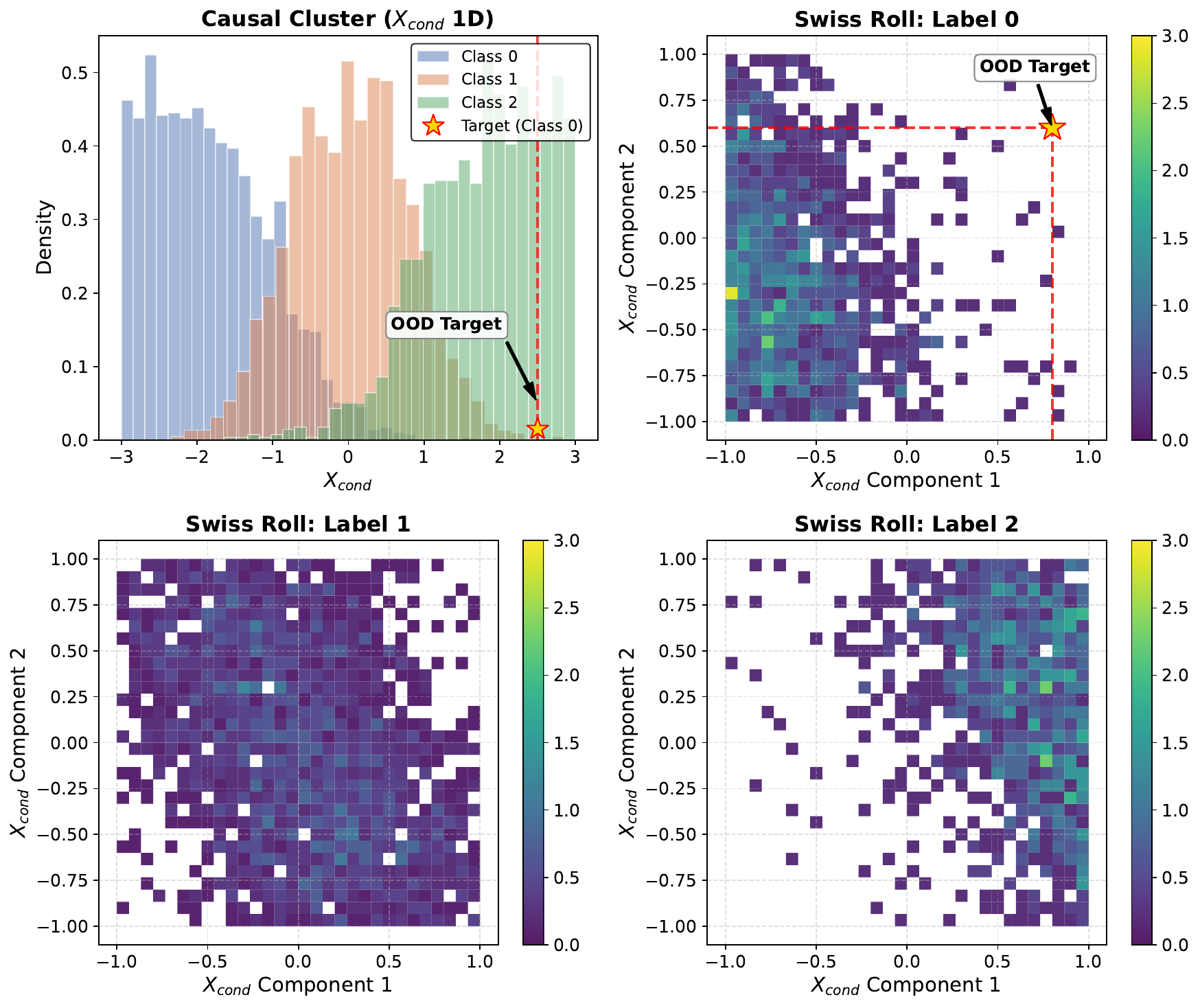} 
\vspace{-10pt}
\caption{\textbf{Spurious correlation and OOD targets in the evaluation 
 datasets.} 
\textbf{(Top Left)} The marginal distributions of $X_{\rm cond}$ given each class label in the Causal Cluster dataset. The target condition for Class 0 (red star) falls into a zero-density region. 
\textbf{(Top Right)} The 2D density plot of $X_{\rm cond}$ for Label 0 in the Swiss Roll dataset. The target condition (red star) is situated outside the observed high-density areas. 
\textbf{(Bottom Row)} The 2D density plots of $X_{\rm cond}$ for Label 1 (left) and Label 2 (right) in the Swiss Roll dataset, illustrating the conditional distributions for the remaining classes. 
In both scenarios, the out-of-distribution (OOD) targets serve as an evaluation of generation capabilities.
}
 \label{artfi_obs}
\vspace{-10pt}
\end{center}
\end{figure}

Using these generation mechanisms, we constructed datasets of 5,000 
samples each. To evaluate model robustness, we generated data using five 
different random seeds and performed five independent training runs for 
each resulting dataset. For the evaluation of the trained models, we 
targeted specific OOD coordinates derived from the data generation 
mechanism, where the combination of $X_{\rm cond}$ and $X_{\rm label}$ is 
extremely rare (marked with stars in Fig.~\ref{artfi_obs}). 
To measure the generation fidelity quantitatively, we drew 300 samples 
from the true distribution and generated 300 empirical samples from each 
trained model at these specific points, subsequently calculating the 
Sliced Wasserstein Distance (SWD) between them. 

\subsubsection{Baselines}
We compare our method against three baselines: 
(1) \textbf{\rm base}, a standard conditional diffusion model trained on the 
empirical distribution without reweighting or guidance;  
(2) \textbf{CFG}, which applies Classifier-Free Guidance 
\cite{ho2021classifierfree} ($\omega=2$, 10\% condition dropout) to the 
Base model to balance fidelity and diversity;  
and (3) \textbf{WDDPM}, a Weighted Diffusion model optimized via 
$\mathcal{L}_{\text{weight}}$ using the exact same density ratio 
estimator $\hat{w}(X)$ used in our DR-Diff. 

\subsubsection{Implementation Details} 
We adopted a Multi-Layer Perceptron (MLP) as the denoising neural 
network for the diffusion model. This is based on findings from tabular 
data generation models, such as TabDDPM\cite{10.5555/3618408.3619133} 
and TabSyn\cite{zhang2024mixedtype}, which indicate that 
MLP architectures are sufficient for such tasks. To strongly condition 
the generation on $X$, we integrated FiLM blocks 
\cite{FiLM} into each MLP layer, which apply an 
affine transformation (scaling $\gamma$ and shifting $\beta$) using the 
encoded information of $X$.

To estimate the weight $w(X)$
, we utilized the LightGBM\cite{10.5555/3294996.3295074}.
 Furthermore, establishing our pseudo-DR target 
 $\tilde{\epsilon}_{\text{\rm DR}} = \hat{\epsilon}_{\text{\rm base}} + \eta 
 \hat{w}(X) (\epsilon - \hat{\epsilon}_{\text{\rm base}})$ requires a step 
 size of $\eta=1$ to match the optimal functional gradient direction. 
 Theorem \ref{theo_riskreduction} mathematically dictates that the loss reduction 
 is guaranteed only if $\eta \max \hat{w}(X) < 2$. Combining these two 
 theoretical requirements necessitates a strict upper bound of $\max 
 \hat{w}(X) < 2$. To satisfy this condition, we clipped the estimated 
 weights during training such that $\max \hat{w}(X) = 1.9$. 
 Hyperparameters are detailed in Appendix \ref{model_conf}. 

\subsubsection{Evaluation Results}
\begin{figure}[tb]
\vspace{-5pt}
\begin{center}
 \begin{tabular}[tb]{c}
 Causal Cluster \\
 \begin{minipage}{0.27\textwidth}
  \includegraphics[width=\linewidth]{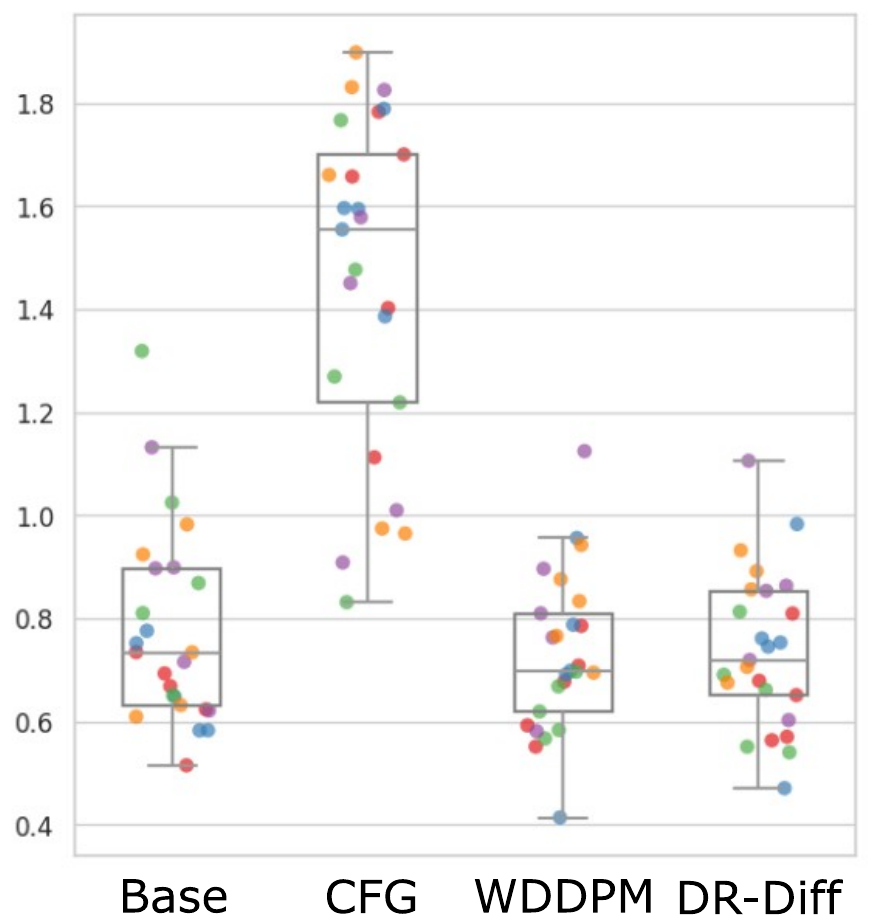} %
 \end{minipage}
\\
\\
 Swiss Roll\\
 \begin{minipage}{0.27\textwidth}
  \includegraphics[width=\linewidth]{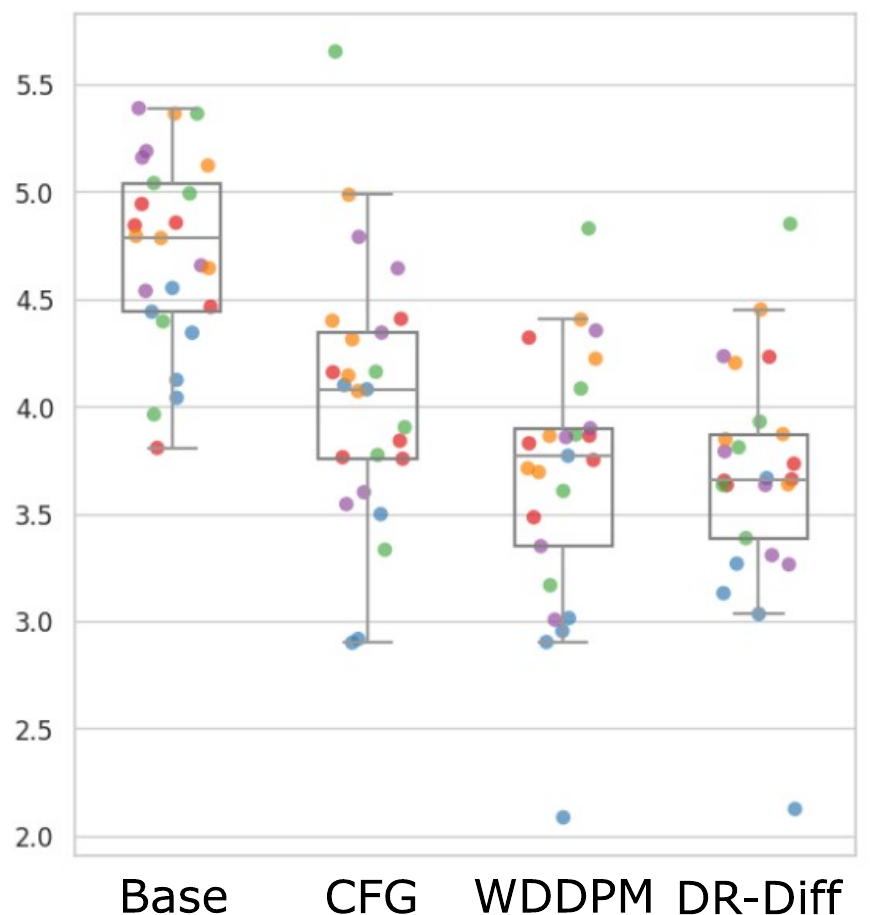} %
 \end{minipage}
\\
\end{tabular}
\caption{\textbf{Quantitative comparison of SWD across methods for OOD targets.} 
The vertical axis represents the Sliced Wasserstein Distance (SWD) 
 between the true distribution and the empirical distribution obtained 
 from each method under OOD target conditions. The 
 color of each plotted point corresponds to results from models trained 
 on datasets generated with the same seed value. Notably, the proposed 
 method achieves high generation performance comparable to WDDPM through 
 only additional training from the Base model. }
 \label{artif1}
\end{center}
\vspace{-15pt}
\end{figure}

\begin{table}[tb]
\caption{
\textbf{Quantitative comparison of the SWD  across methods for OOD targets.} 
(Values reported as Q1 / Median / Q3).
}
\label{artif1_tab}
\vspace{-15pt}
 \begin{center}
\begin{tabular}{lcc}
\toprule
\textbf{Method} & \textbf{Causal Cluster} & \textbf{Swiss Roll} \\
\midrule
Base model & 0.632 / 0.734 / 0.897 & 4.442 / 4.785 / 5.041 \\
CFG & 1.219 / 1.555 / 1.700 & 3.756 / 4.080 / 4.344 \\
WDDPM & 0.619 / 0.698 / 0.810 & 3.349 / 3.770 / 3.898 \\
DR-Diff(Ours) & 0.651 / 0.719 / 0.854 & 3.386 / 3.661 / 3.871 \\
\bottomrule
\end{tabular}\end{center}
\vspace{-15pt}
\end{table}

\begin{figure*}[tb]
\begin{center}
  \includegraphics[width=0.85\linewidth]{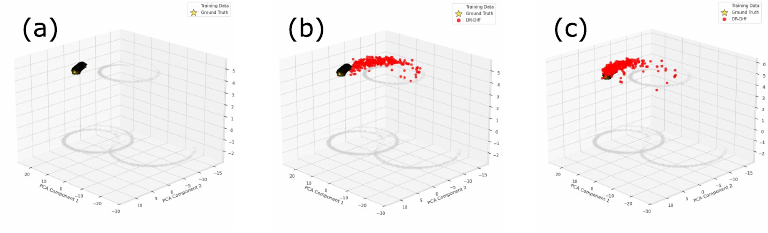}
\caption{\textbf{Visualization of generated samples for the Swiss Roll dataset.} 
To visualize the underlying data manifold structure, the original 
 $\mathbb{R}^{10}$ space is projected down to 3 dimensions via PCA, upon 
 which the generated points are plotted. (a) The light gray points 
 denote the training data, while the stars 
 represent the 
 ground-truth points generated under the current OOD condition. (b) 
 Generation results from a model yielding a slightly poor SWD of 
 3.7. (c) Generation results from a model achieving an SWD of 2.12. Both 
 models successfully recover the global roll structure; however, it can 
 be observed that the large SWD values stem from the high variance in 
 estimating the angular velocity of the roll. 
}
 \label{artf_swiss}
\end{center}
\vspace{-5pt}
\end{figure*}

Fig.~\ref{artif1} and Table \ref{artif1_tab} summarize the SWD across all runs. 
When comparing the medians in the Causal Cluster dataset, the Base model 
already demonstrates good predictive performance at the initial stage, 
and both the proposed method and WDDPM maintain these results. 
Conversely, CFG exhibits a bias that pulls the generation toward 
high-density regions of the observed data as the conditioning strength 
increases. Consequently, it is heavily influenced by spurious 
correlations within the training data, which suggests a significant 
deterioration in predictive accuracy for OOD targets. 

Regarding the results for the Swiss Roll dataset, the SWD of the Base 
model is generally high because the underlying standard diffusion model 
fails to acquire feature representations in the OOD regions 
of the complex manifold structure. However, CFG, WDDPM, and 
the proposed method all significantly improve the SWD compared to the 
Base model. Notably, WDDPM and the proposed method outperform CFG, both 
demonstrating similarly high generation capabilities. 

Furthermore, the overall SWD values tend to be larger for the Swiss Roll 
than for the Causal Cluster. This is likely due to the inherent 
difficulty of learning the non-linear roll structure of the Swiss Roll 
and extrapolating into unknown regions. 
Indeed, as shown in Fig.~\ref{artf_swiss}, which compares 
the data generated by a model with an SWD of 3.73 
against one with an SWD of 2.12, there is substantial variance in the 
manifold extrapolation within the OOD region. This visually confirms the 
high uncertainty in generation for the target condition. 

Computationally, 
as detailed in Appendix \ref{model_conf},
whereas baselines required 100?200 epochs of training 
from scratch, DR-Diff achieved these results with 40 epochs of 
additional training on the Base model, enabling efficient target 
adaptation without prohibitive retraining. 

In conclusion, we demonstrated that for generating OOD targets, the 
proposed method achieves high performance equivalent to an approach 
trained from scratch using a specialized loss function (WDDPM), solely 
through additional training on an existing Base model. This is highly 
advantageous as it eliminates the influence of spurious correlations 
while keeping computational costs low. 

\subsection{Ablation Study: Impact of Accurate Weight Estimation}
Based on Theorem \ref{theo_riskreduction}, the expected loss of the noise prediction error 
under the target distribution is guaranteed to decrease as long as the 
estimated weight satisfies $\hat{w} < 2$. This implies 
that any weight estimation within this bounded range might suffice. To 
investigate whether accurately estimating $W$ is strictly necessary in 
practice, we conduct an ablation study. 

\subsubsection{Setup and Results:}

\begin{figure}[tb]
\begin{center}
 \begin{tabular}[tb]{c}
 Causal Cluster \\
 \begin{minipage}{0.27\textwidth}
  \includegraphics[width=\linewidth]{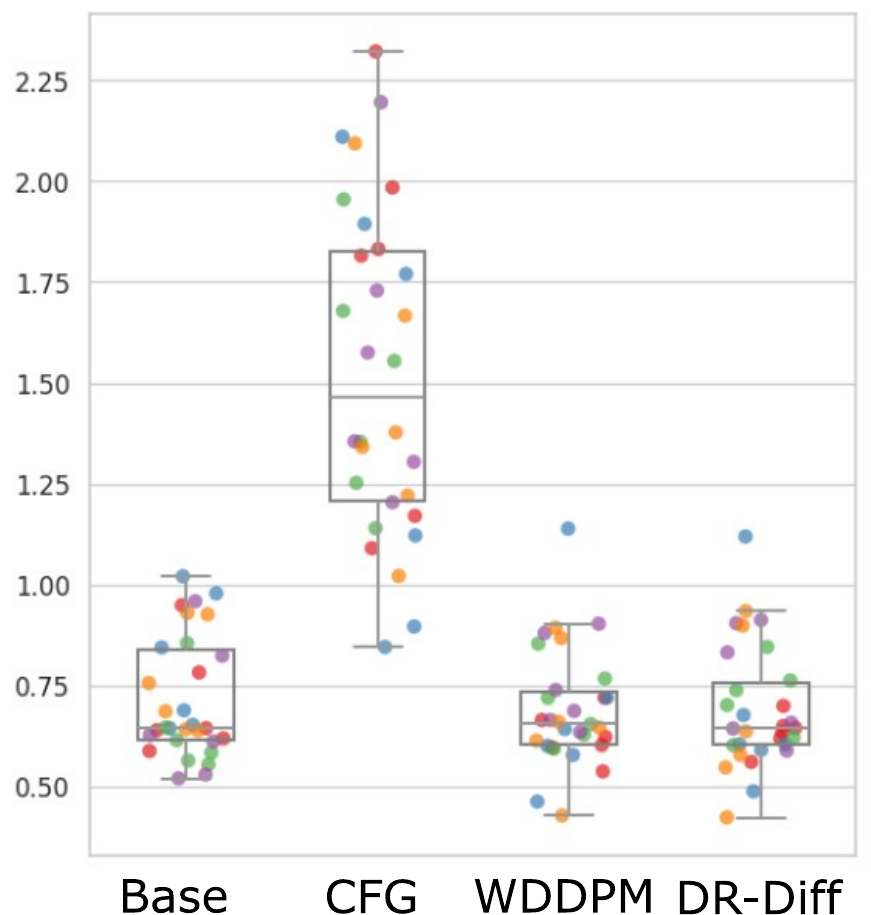} %
 \end{minipage}
\\
\\
 Swiss Roll\\
 \begin{minipage}{0.27\textwidth}
  \includegraphics[width=\linewidth]{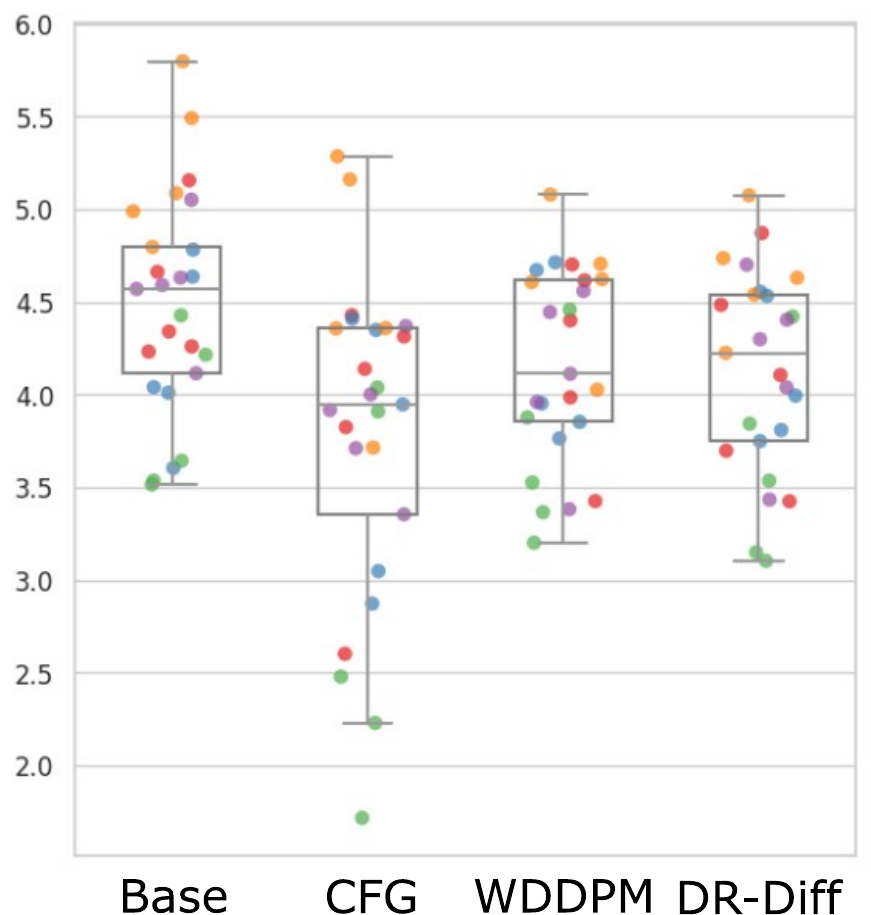} %
 \end{minipage}
\\
\end{tabular}
\caption{\textbf{Ablation study evaluating the impact of accurate weight estimation.} 
The vertical axis shows the SWD between the true and empirical 
 distributions for OOD targets. As an ablation, we present the results 
 of replacing the properly estimated $w$ with random weights $w \sim U(0, 
 1.9)$. It is observed that destroying the structure of $w$ limits the 
 improvement in generation quality, preventing it from surpassing the 
 performance of CFG. This confirms that while bounded random weights are 
 theoretically safe, accurately estimating $w$ is critical for the 
 optimal estimation of the noise prediction function.} 
 \label{artif1_abulation}
\end{center}
\vspace{-15pt}
\end{figure}

\begin{table}[tb]
\caption{
\textbf{Quantitative comparison of the SWD 
 across methods for OOD targets under the ablation setup.} 
}
\label{artif1_abulation_tab}
\vspace{-15pt}
 \begin{center}
\begin{tabular}{lcc}
\toprule
\textbf{Method} & \textbf{Causal Cluster} & \textbf{Swiss Roll} \\
\midrule
  Base model&  0.616  / 0.645  / 0.840& 4.116  / 4.570  / 4.796\\
   CFG      &  1.209  / 1.466  / 1.828& 3.355  / 3.947  / 4.357\\
   WDDPM    &  0.605  / 0.658  / 0.734& 3.853  / 4.113  / 4.616\\
   DR-Diff(Ours)&  0.603  / 0.644  / 0.757& 3.749  / 4.225  / 4.539\\
\bottomrule
\end{tabular}\end{center}
\vspace{-15pt}
\end{table}

We simulate a flawed estimation scenario by intentionally destroying the 
structure of $W$. Specifically, we replace the accurately estimated 
weights with random values sampled from a uniform distribution within 
the theoretically safe bounded range, $W \sim U(0, 1.9)$. Following the 
same procedure as in Section 4.1, we apply this setup to both the Causal 
Cluster and Swiss Roll datasets. The results are summarized in 
Fig.~\ref{artif1_abulation} and Table \ref{artif1_abulation_tab}.

Under the theoretical formulation of doubly robust estimators, if the 
base model $\epsilon_{\rm base}$ perfectly captures the true noise 
prediction, optimal generation is guaranteed regardless of the accuracy 
of the structural weight $W$. Indeed, in the Causal Cluster dataset, 
where $\epsilon_{\rm base}$ already estimates the distribution with high 
accuracy (as observed in Section 4.1), the model with randomly assigned 
weights successfully maintains performance comparable to the Base model. 
Conversely, in the Swiss Roll dataset, the initial estimation by 
$\epsilon_{\rm base}$ is suboptimal. In this scenario, although the random 
weight model still improves upon the Base model (due to the $\hat{w} < 
2$ condition), its degree of improvement is lower than that of CFG and 
falls significantly short of the proposed method with accurately 
estimated weights. 

\subsection{Fairness via Mitigating Spurious Dependence on Real-World Data.}
Through evaluations on synthetic data, we demonstrated that our proposed 
method achieves generation performance equivalent to the idealized WDDPM 
under the target distribution. Building on this, this section 
investigates its practical utility in real-world scenarios, evaluating 
whether data synthesized by our method can effectively mitigate 
structural biases to train fair downstream classifiers. 
Due to the substantial computational cost of training diffusion models on real-world datasets, we report the results on a single run.

\subsubsection{Datasets and Preprocessing}
We utilize two real-world datasets, LSAC and Jigsaw, which exhibit 
strong spurious correlations between sensitive attributes and target 
outcomes (details in Appendix \ref{LSAC}). The LSAC task predicts bar 
exam passage based on law school academic performance and demographic 
attributes (e.g., race). The Jigsaw task determines the toxicity of 
texts annotated for mentions of specific identity attributes (e.g., race 
or gender). Due to the high diversity and computational cost of directly 
handling text, we applied a Sentence Transformer (nomic-embed-text-v1.5) 
to compress the texts into 768-dimensional embedding vectors, enabling 
scalable and modality-agnostic generation. 

\subsubsection{Evaluation Protocol}
For both datasets, we split the data into 80/20 train-test sets and 
trained a baseline MLP classifier on the 
training set (hereafter, the Original Model). Inheriting the empirical 
distribution, the Original Model typically exhibits significant False 
Positive Rate (FPR) disparities across specific attributes (e.g., race 
in LSAC) due to spurious correlations. We evaluate whether fine-tuning 
the Original Model on augmented data with a reduced learning rate can 
yield a fairer classifier. Models are evaluated by comparing the overall 
and subgroup-specific Accuracy, TPR, and FPR on the test set.
Hyperparameters are detailed in Appendix \ref{model_conf}. 

\subsubsection{Baselines and Proposed Method}
To mitigate structural biases, we compare three approaches:
\begin{itemize}
\item \textbf{Reweighing Method(RW):} Adjusts sample weights during the MLP 
      loss calculation using the Reweighing algorithm from the AI 
      Fairness 360 (AIF360) package \cite{10.1007/s10115-011-0463-8}. 
\item \textbf{Oversampling Method(OS):} To focus on instances misclassified 
      due to inherent biases, we first perform 5-fold cross-validation 
      on the training set to identify misclassified data points. 
      These points are simply duplicated by a factor of $N$ 
      ($N=4$ in our experiments) and appended to the training data. 
\item \textbf{Proposed Method(Ours):} Targeting the same points identified 
      above, we use our generative model to generate $2N$ new data 
      points for each instance while retaining the original condition 
      (e.g., sensitive attributes): $N$ points with the original  
      label, and $N$ points with the inverted label (e.g., 
      generating $N$ ``failed'' and $N$ ``passed'' instances for those 
      misclassified in LSAC).
\end{itemize}

\subsubsection{Results}
\begin{table}[tb]
\begin{center}
\caption{Performance metrics of the classifiers on the LSAC dataset 
 across racial subgroups. 'Gap' indicates the absolute difference 
 between Group 0 (White) and Group 1 (Non-White) defined by the race 
 attribute.} 
 \label{resultab:lsac}
\begin{tabular}{llcccc}
\toprule
\textbf{Metric} & \textbf{Group} & \textbf{Original} & \textbf{OS} & \textbf{RW} & \textbf{Ours} \\
\midrule
\multirow{4}{*}{Accuracy} &All&0.911&0.902&0.902&0.907\\
&Group0&0.921&0.910&0.922&0.916\\
&Group1&0.755&0.777&0.592&0.777\\
&Gap&0.166&0.133&0.330&0.139\\
\midrule
\multirow{4}{*}{FPR}&All&0.763&0.559&0.883&0.635\\
&Group0&0.897&0.638&0.845&0.749\\
&Group1&0.385&0.333&0.990&0.313\\
&Gap&0.511&0.305&0.145&0.437\\
\midrule
\multirow{4}{*}{TPR} &All&0.984&0.952&0.987&0.966\\
&Group0&0.990&0.956&0.987&0.972\\
&Group1&0.854&0.854&1.000&0.839\\
&Gap&0.136&0.102&0.013&0.132\\
\bottomrule
 \end{tabular}
\vspace{-15pt}
\end{center} 
\end{table}
\vspace{-5pt}
\begin{table}[tb]
 \begin{center}
\caption{Performance metrics of the classifiers on the Jigsaw dataset 
  (Target attribute: mentions of the 'Black' identity). Group 1 
  represents texts containing mentions of the identity, while Group 0 
  represents texts without. 'Gap' indicates the absolute difference 
  between the subgroups.} 
 \label{resultab:jigsaw_black}
\begin{tabular}{llcccc}
\toprule
\textbf{Metric} & \textbf{Group} & \textbf{Original} & \textbf{OS} & \textbf{RW} & \textbf{Ours} \\
\midrule
\multirow{4}{*}{Accuracy}  
&All&0.910&0.905&0.908&0.908\\
&Group0&0.915&0.911&0.914&0.913\\
&Group1&0.758&0.753&0.742&0.762\\
&Gap&0.158&0.158&0.173&0.152\\
\midrule
\multirow{4}{*}{FPR} 
&All&0.029&0.043&0.029&0.030\\
&Group0&0.027&0.039&0.029&0.028\\
&Group1&0.108&0.158&0.031&0.074\\
&Gap&0.082&0.119&0.003&0.046\\
\midrule
\multirow{4}{*}{TPR} 
&All&0.430&0.499&0.413&0.417\\
&Group0&0.426&0.492&0.432&0.419\\
&Group1&0.465&0.558&0.247&0.403\\
&Gap&0.039&0.066&0.185&0.015\\
\bottomrule
  \end{tabular}
\vspace{-15pt}
 \end{center}
\end{table}

\begin{table}[tb]
 \begin{center}
\caption{Performance metrics of the classifiers on the Jigsaw dataset 
  (Target attribute: mentions of the 'Male' identity). Group 1 
  represents texts containing mentions of the identity, while Group 0 
  represents texts without. 'Gap' indicates the absolute difference 
  between the subgroups.} 
 \label{resultab:jigsaw_male}
\begin{tabular}{llcccc}
\toprule
\textbf{Metric} & \textbf{Group} & \textbf{Original} & \textbf{OS} & \textbf{RW} & \textbf{Ours} \\
\midrule
\multirow{4}{*}{Accuracy}  
&All&0.910&0.905&0.908&0.908\\
&Group0&0.913&0.910&0.912&0.912\\
&Group1&0.878&0.869&0.876&0.872\\
&Gap&0.035&0.041&0.036&0.040\\
\midrule
\multirow{4}{*}{FPR} 
&All&0.029&0.043&0.036&0.030\\
&Group0&0.028&0.040&0.035&0.028\\
&Group1&0.042&0.070&0.045&0.044\\
&Gap&0.014&0.031&0.010&0.017\\
\midrule
\multirow{4}{*}{TPR} 
&All&0.430&0.499&0.468&0.417\\
&Group0&0.429&0.493&0.473&0.418\\
&Group1&0.437&0.533&0.436&0.415\\
&Gap&0.008&0.040&0.037&0.003\\
\bottomrule
  \end{tabular}
\vspace{-15pt}
 \end{center}
\end{table}

The evaluation results for LSAC are shown in Table \ref{resultab:lsac}, 
and those for Jigsaw are shown in Tables \ref{resultab:jigsaw_black} and 
\ref{resultab:jigsaw_male}. Although the Jigsaw dataset contains 
multiple attribute annotations, we present two representative cases due 
to space constraints: mentions of the ``Black'' identity, where the 
initial model exhibits significant disparities (Gap) in FPR/TPR (Table 
\ref{resultab:jigsaw_black}), and mentions of the ``Male'' identity, 
where the initial disparities are relatively small (Table 
\ref{resultab:jigsaw_male}). 

As shown in Tables \ref{resultab:lsac} and \ref{resultab:jigsaw_black}, 
the Original exhibits large FPR gaps between subgroups. Our proposed 
method successfully reduces both the FPR for Group 1 and the overall FPR 
gap, while maintaining the smallest drop in overall accuracy among all 
compared methods. The RW method substantially reduces the FPR gap in all 
cases. However, this introduces notable performance trade-offs. In Table 
\ref{resultab:lsac}, the overall FPR increases significantly. In Table 
\ref{resultab:jigsaw_black}, while RW reduces the FPR for Group 1, it 
simultaneously causes a substantial decrease in TPR, indicating a 
negative impact on the classifier's overall utility. The OS method 
exhibits inconsistent behavior depending on the dataset, and in the 
Jigsaw task, it even worsens the FPR and its corresponding gap. 
Furthermore, as demonstrated in Table \ref{resultab:jigsaw_male}, in 
scenarios where the baseline FPR gap is already small, OS degrades 
accuracy while increasing the FPR. In contrast, both our proposed method 
and RW function safely without undermining the performance of the 
original classifier. 

While this study focused on comparing DR-Diff with standard augmentation 
techniques (OS, RW), isolating the generative benefits by comparing with 
Base, CFG, and WDDPM-based augmentations remains an important future 
direction. However, since DR-Diff matches the distribution quality of 
WDDPM (as shown in Section \ref{synthetic_exp}), we theoretically anticipate it would 
yield a similar fairness-utility trade-off to WDDPM-based augmentation, 
but at a fraction of the computational cost. 

Overall, these results show that augmenting misclassified instances with 
condition-retained, label-flipped data robustly improves the 
fairness-utility trade-off without sacrificing accuracy. 

\section{Conclusion.}
Conditional diffusion models memorize spurious correlations within 
training data. Consequently, they deviate from the ideal target 
distribution. To resolve this problem, minimizing a covariate 
shift-adjusted loss function (target risk) is effective. However, 
retraining models from scratch for every covariate shift is 
computationally inefficient.  

To address this cost bottleneck, we proposed DR-Diff, which tackles this 
issue by fine-tuning existing base models. Specifically, we derive a 
pseudo-target based on the functional gradient of the target risk and 
train the noise prediction function to predict it. 
Theoretically, we demonstrated that the resulting noise prediction 
function is guaranteed to reduce the target risk at the population level
(Theorem \ref{theo_riskreduction}) and 
achieves pseudo-Double Robustness. The latter provides a degree of 
resilience against both estimation errors in the density ratio model and 
misspecifications of the base model. 

Through numerical experiments on synthetic datasets, we confirmed that 
our method achieves noise prediction performance comparable to 
performing weighted optimization from scratch. Furthermore, by applying 
it to real-world datasets, we demonstrated its ability to significantly 
reduce inter-group disparities in downstream classifiers. 

Our current evaluations primarily focus on tabular and embedding data. 
Nevertheless, this approach is fundamentally model-agnostic. Therefore, 
it holds great promise for future extensions to image and multimodal 
generation.

\appendices
\section{Proofs of Proposition \ref{prop_neyman_orthogonality}.}
\label{proof:neyman_orthogonality}

In this section, we provide the formal proof for 
Proposition \ref{prop_neyman_orthogonality}.
First, we show the definition of Neyman Orthogonality.
We adopt the formulation and notation provided by 
\cite{OrthogonalStatisticalLearning}.

For i.i.d. observed data $D$, let 
$\theta$ be the target parameter to be estimated and $L_D$ be the loss 
function used for optimization. Here, $L_D$ depends not only on $\theta$ 
but also on a nuisance parameter (function) $g$, whose true function 
$g_0$ is unknown. Neyman orthogonality refers to the condition where the 
cross-derivative of $L_D$ with respect to the directional derivatives of 
$g$ and $\theta$ vanishes at the truth. The rigorous definitions are as 
follows: 

\begin{definition}[Directional Derivative]
\label{def:DirectionalDerivative}
Let $\mathcal{F}$ be a vector space of functions. For a functional $F : \mathcal{F} \to \mathbb{R}$, we define the derivative operator $D_f F(f)[h] = \partial_t F(f + th)|_{t=0}$ for a pair of functions $f, h \in \mathcal{F}$.
\end{definition}

\begin{definition}[Neyman Orthogonal]
\label{def:NeymanOrthogonal}
We say $L_D$ is Neyman orthogonal when the following holds:
$$\forall \theta \in \Theta, \forall g \in \mathcal{G}, \quad D_g D_\theta L_D(\theta^*, g_0) [\theta - \theta^*, g - g_0] = 0.$$
where $\theta^*$ and $g_0$ represent the true parameter and function, respectively.
\end{definition}

We prove that the weighted objective function $\mathcal{L}_{\text{\rm weighted}}(\epsilon_\theta, w)$ satisfies Neyman orthogonality according to Definition \ref{def:NeymanOrthogonal}.

First, we verify the well-definedness of the directional derivatives. Since the space of noise prediction functions $\mathcal{F}$ and the space of weight functions $\mathcal{W}$ are both convex sets, the convex combinations $(1-t)\epsilon_\theta + t\epsilon_\theta'$ and $(1-t)w + t w'$ for arbitrary $\epsilon_\theta, \epsilon_\theta' \in \mathcal{F}$ and $w, w' \in \mathcal{W}$ belong to their respective spaces for $t \in [0, 1]$. Therefore, the directional derivatives are well-defined.

Let $\epsilon_{\theta^*} := \mathbb{E}[\epsilon \mid y_t, t, X]$ be the true noise prediction function and $w^*$ be the true weight function. We denote their perturbations as $\Delta \epsilon := \epsilon_\theta - \epsilon_{\theta^*}$ and $\Delta w := w - w^*$, respectively.
First, we compute the directional derivative of $\mathcal{L}_{\text{\rm weighted}}$ with respect to $\epsilon_\theta$ evaluated at $\epsilon_{\theta^*}$:
\begin{align}
D_{\epsilon_\theta} \mathcal{L}_{\text{\rm weighted}}(\epsilon_{\theta^*}, w) [\Delta \epsilon] 
\!&= \!\left. \frac{d}{dt} \mathbb{E}\! \left[ w(X) \|\epsilon\!-\! (\epsilon_{\theta^*}\!+\! t \Delta \epsilon)\|^2 \right] \right|_{t=0} \nonumber \\
&=\! \mathbb{E} \left[ -2 w(X) (\epsilon - \epsilon_{\theta^*})^T \Delta \epsilon \right].
\end{align}
Next, we compute the directional derivative of the obtained functional gradient with respect to $w$ evaluated at $w^*$:
\begin{align}
&D_w D_{\epsilon_\theta} \mathcal{L}_{\text{\rm weighted}}(\epsilon_{\theta^*}, w^*) [\Delta \epsilon, \Delta w]\nonumber\\ 
&= \left. \frac{d}{dt} \mathbb{E} \left[ -2 (w^*(X) + t \Delta w(X)) (\epsilon - \epsilon_{\theta^*})^T \Delta \epsilon \right] \right|_{t=0} \nonumber \\
&= \mathbb{E} \left[ -2 \Delta w(X) (\epsilon - \epsilon_{\theta^*})^T \Delta \epsilon \right].
\end{align}
Here, we apply the law of total expectation by conditioning on $(y_t, t, X)$. From the definition of the true noise prediction function, we have $\mathbb{E}[\epsilon \mid y_t, t, X] = \epsilon_{\theta^*}$, which directly implies $\mathbb{E}[\epsilon - \epsilon_{\theta^*} \mid y_t, t, X] = 0$. Consequently, we obtain:
\begin{align}
& D_w D_{\epsilon_\theta} \mathcal{L}_{\text{\rm weighted}}(\epsilon_{\theta^*}, w^*) [\Delta \epsilon, \Delta w] \nonumber\\
&= \mathbb{E}_{y_t, t, X} \left[ -2 \Delta w(X) \cdot \mathbb{E}[\epsilon - \epsilon_{\theta^*} \mid y_t, t, X]^T \Delta \epsilon \right] 
= 0.\nonumber
\end{align}%
Therefore, according to Definition \ref{def:NeymanOrthogonal}, $\mathcal{L}_{\text{\rm weighted}}(\epsilon_\theta, w)$ satisfies Neyman orthogonality at $\epsilon_{\theta^*}$ and $w^*$. 

\section{Datasets and Preprocessing.}
\subsection{Synthetic Data Generation.}
\label{syntheticgen}
\textbf{Causal Clusters: }

We designed a synthetic dataset to evaluate the model's ability to recover the conditional distribution 
$P(y_0 | X_{\rm label}, X_{\rm cond})$ under strong confounding. 
The generation process for $N$ samples is defined as follows:
\begin{enumerate}
 \item Exogenous Confounder ($X_{\rm cond}$): 
The continuous conditioning variable is sampled from a uniform distribution:
$X_{cond, i} \sim \text{Uniform}(-3, 3).$
 \item Causal Mechanism $X_{\rm cond} \to X_{\rm label}$: 
The discrete class label $X_{label, i} \in \{0, 1, 2\}$ is sampled 
according to probabilities derived from $X_{\rm cond}$-dependent logits:
$$P(X_{label, i} = k | X_{cond, i}) = \frac{\exp(\text{logit}_{ik})}{\sum_{j=0}^2 \exp(\text{logit}_{ij})}.$$
where the logits create non-linear class imbalances based on 
the value of the confounder:
$\text{logit}_{i0} = -2.0 X_{cond, i}$, 
$\text{logit}_{i1} = -0.5 X_{cond, i}^2 + 2$, 
$\text{logit}_{i2} = 2.0 X_{cond, i}.$
\item Outcome Generation $X_{\rm label}, X_{\rm cond} \to y_0$: 
The observed 2D target $y_{0, i} \in \mathbb{R}^2$ is generated by 
shifting class-specific centroids $\mu_{X_{label, i}}$ with a non-linear function of the confounder:
$y_{0, i} = \mu_{X_{label, i}} + \text{shift}(X_{cond, i}) + \xi_i, \quad \xi_i \sim \mathcal{N}(0, 0.4^2 \mathbf{I}).$
The centroids are defined as 
$\mu_0 = (-3, -2)^\top, \mu_1 = (0, 3)^\top, \mu_2 = (3, -2)^\top$ 
and the shift function is $\text{shift}(C) = (C, \sin(2C))^\top$.
\end{enumerate}

\textbf{Swiss Roll (10D)):}

We designed a synthetic dataset to evaluate the model's ability to recover the conditional distribution $P(y_0 | X_{\rm label}, X_{\rm cond})$ under complex geometric confounding. The generation process for $N$ samples is defined as follows:
\begin{enumerate}
 \item Exogenous Confounder ($X_{\rm cond}$): 
The continuous conditioning vector is sampled uniformly from a $d_c$-dimensional hypercube:
$X_{cond, i} \sim \text{Uniform}(-1, 1)^{d_c}.$
 \item Causal Mechanism $X_{\rm cond} \to X_{\rm label}$: 
The discrete class label $X_{label, i} \in \{0, 1, 2\}$ is sampled according to probabilities derived from $X_{\rm cond}$-dependent logits:
$$P(X_{label, i} = k | X_{cond, i}) = \frac{\exp(\text{logit}_{ik})}{\sum_{j=0}^2 \exp(\text{logit}_{ij})}$$
where the logits create non-linear class distributions based on 
the aggregated confounding effect 
$e_{i} = \text{clip}(\sqrt{d_c} X_{cond, i}^\top \mathbf{w}, -1.0, 1.0)$
 with a decaying weight vector $\mathbf{w}= [2/3, 1/3]^\top$ (for $d_c = 2$):
$\text{logit}_{i0} = -3.0 e_{i}$, 
$\text{logit}_{i1} = 3.0 (1.0 - |e_{i}|)$, $\text{logit}_{i2} = 3.0 e_{i}.$
\item Outcome Generation $X_{\rm label}, X_{\rm cond} \to y_0$: 
The observed high-dimensional target $y_{0, i} \in \mathbb{R}^{d_x}$ is generated 
by embedding a 6D latent representation 
$\mathbf{h}_i = [\mathbf{v}_i^\top, \mathbf{m}_{X_{label, i}}^\top]^\top$. 
The 3D volumetric coordinates $\mathbf{v}_i \in \mathbb{R}^3$ construct 
class-conditional Swiss rolls:
$$\mathbf{v}_i = \begin{pmatrix} \theta_i \cos(\omega_{X_{label, i}} \theta_i) + \Delta x_{X_{label, i}} \\ X_{cond, i}^{(d_c)} \\ \theta_i \sin(\omega_{X_{label, i}} \theta_i) + \Delta z_{X_{label, i}} \end{pmatrix} + \mathbf{n}_i.$$
Here, the angular progression $\theta_i$ is determined by the mean of $X_{cond, i}$ 
with added uniform noise drawn from $\text{Uniform}(-0.05, 0.05)$.
The parameters $\omega_k \in \{1.0, 1.5, 0.7\}$ and $(\Delta x_k, \Delta z_k)$ 
define the class-specific unrolling speed and spatial shifts, respectively. 
Volume is added via isotropic noise $\mathbf{n}_i$ sampled uniformly from a 3D ball. 
The margin vector $\mathbf{m}_k \in \mathbb{R}^3$ enforces orthogonal class separability ($5.0$ at index $k$, $0$ otherwise). Finally, $\mathbf{h}_i$ is projected into the target space via a random orthogonal matrix $P \in \mathbb{R}^{6 \times d_x}$:
$y_{0, i} = P^\top \mathbf{h}_i.$
\end{enumerate}

\subsection{Real-World Datasets}\label{LSAC}
All datasets used in our experiments are publicly available. 
The stats of datasets are shown in Table. \ref{tab:dataset_summary}.
LSAC is sourced from the Law School Admission Council National Longitudinal Bar Passage Study 
(\url{https://www.kaggle.com/datasets/danofer/law-school-admissions-bar-passage}).
Jigsaw is sourced from Jigsaw Unintended Bias in Toxicity Classification
(\url{https://www.kaggle.com/c/jigsaw-unintended-bias-in-toxicity-classification}).
For the Jigsaw dataset, we preprocessed the data by extracting only the 
annotated records and selecting five specific annotation categories for 
our experiments. 

\begin{table*}[tb]
\centering
\small 
\caption{Summary and Specification of Real-World Datasets}
\label{tab:dataset_summary}
\begin{tabular}{lrrll}
\toprule
\textbf{Dataset} & \textbf{\# Samp.} & \textbf{\# Feat.} & \textbf{Target ($X_{\rm label}$)} & \textbf{Auxiliary Attributes ($X_{\rm cond}$)} \\ 
\midrule
LSAC & 21,791 & 12 & Bar exam passage & 
Race, Gender, LSAT score, Undergraduate GPA, Family income level\\
Jigsaw& 405,130& 5& Toxicity of text &  Mentioned demographic identities (e.g., Black, White, Male, Female, LGBTQ)\\  
\bottomrule
\end{tabular}
\vspace{-10pt}
\end{table*}

\section{Model Configurations}\label{model_conf}
To ensure the full reproducibility of our experiments, we provide the 
detailed hyperparameters and architectural configurations used in our 
numerical experiments in Table \ref{tab:hyperparameters}. 

In real-world datasets experiments, 
for the classification models, we employ a 3-layer MLP for both datasets, 
incorporating Batch Normalization, activation functions, and Dropout. 
Specifically, the LSAC model uses hidden dimensions of 128 and 64, 
LeakyReLU (negative slope 0.1), and a dropout rate of 0.2. The Jigsaw 
model uses hidden dimensions of 512 and 256, ReLU, and a dropout rate of 
0.3. Both models output a 1-dimensional logit. 

\begin{table*}[htbp]
\centering
\caption{Detailed Hyperparameters and Model Architectures}
\label{tab:hyperparameters}
\small 
\begin{tabular}{l c c c c}
\toprule
\textbf{Hyperparameter / Setting} & \multicolumn{2}{c}{\textbf{Synthetic Dataset }} & \multicolumn{2}{c}{\textbf{Real-World Datasets}} \\
& \textbf{Causal Cluster}&\textbf{Swiss Roll} & \textbf{LSAC}&\textbf{Jigsaw} \\
\midrule
&\multicolumn{4}{c}{\textit{Model Architecture}} \\
\midrule
Base Structure & \multicolumn{4}{c}{FiLM-based 2-layer MLP } \\
Input / Output Dimension & \multicolumn{4}{c}{$\dim(y_0) \rightarrow \dim(y_0)$} \\
Hidden Dimension &[128, 128] & [256, 256] & [128, 128] &[2048, 2048]\\
Label Embedding & 3 classes $\rightarrow$ 128 &3 classes $\rightarrow$ 256 & 2 classes $\rightarrow$ 128 & 2 classes $\rightarrow$ 2048 \\
Conditioning Embedding & $\dim(X_{\rm cond})$ $\rightarrow$ 128 & $\dim(X_{\rm cond})$ $\rightarrow$ 256 & $\dim(X_{\rm cond})$ $\rightarrow$ 128 & $\dim(X_{\rm cond})$ $\rightarrow$ 2048\\
Core Network Activation & \multicolumn{4}{c}{SiLU}\\
Dropout Rate & \multicolumn{4}{c}{$p = 0$} \\
\midrule
&\multicolumn{4}{c}{\textit{Diffusion Process}} \\
\midrule
Total Timesteps ($T$) & \multicolumn{4}{c}{1000} \\
Noise Schedule & \multicolumn{4}{c}{Linear ($\beta_1 = 10^{-4}, \beta_T = 0.02$)} \\
Sampler & \multicolumn{4}{c}{standard reverse process (1000 steps)} \\
Optimizer & \multicolumn{4}{c}{AdamW ($\beta_1=0.9, \beta_2=0.999$, Weight Decay$=0.01$)} \\
Batch Size & 64 &64 &64 &512  \\
Learning Rate ($\gamma$) (Base, DR-Diff) & {5$ \times 10^{-5}$, 1$\times 10^{-5}$}
 &{10$\times 10^{-5}$, 10$\times 10^{-5}$} & 10$\times 10^{-7}$, 6$\times 10^{-7}$ 
 & 50$\times 10^{-5}$, 2$\times 10^{-5}$\\
Training Iterations (Base, DR-Diff) &100, 40& 200, 40 &200, 40&10, 2  \\
\midrule
&\multicolumn{4}{c}{\textit{ Nuisance Estimator}} \\
\midrule
Nuisance Model ($\hat{w}$) & \multicolumn{4}{c}{LightGBM} \\
Nuisance Input Features & \multicolumn{4}{c}{Auxiliary Attributes ($X_{\rm cond}$)} \\
Cross-fitting Folds ($K$) & \multicolumn{4}{c}{5} \\
\bottomrule
\end{tabular}
\end{table*}

\bibliography{references_upload}
\bibliographystyle{IEEEtran}

\end{document}